# Feature-context driven Federated Meta-Learning for Rare Disease Prediction

**Bingyang Chen, Tao Chen, Xingjie Zeng, Weishan Zhang, Qinghua Lu, Zhaoxiang Hou, Jiehan Zhou and Sumi Helal,**
*IEEE Fellow*

**Abstract:** Millions of patients suffer from rare diseases around the world. However, the samples of rare diseases are much smaller than those of common diseases. In addition, due to the sensitivity of medical data, hospitals are usually reluctant to share patient information for data fusion citing privacy concerns. These challenges make it difficult for traditional AI models to extract rare disease features for the purpose of disease prediction. In this paper, we overcome this limitation by proposing a novel approach for rare disease prediction based on federated meta-learning. To improve the prediction accuracy of rare diseases, we design an attention-based meta-learning (ATML) approach which dynamically adjusts the attention to different tasks according to the measured training effect of base learners. Additionally, a dynamic-weight based fusion strategy is proposed to further improve the accuracy of federated learning, which dynamically selects clients based on the accuracy of each local model. Experiments show that with as few as five shots, our approach out-performs the original federated meta-learning algorithm in accuracy and speed. Compared with each hospital's local model, the proposed model's average prediction accuracy increased by 13.28%.

*Index Terms*--Attention-based meta-learning, dynamic-weight based fusion strategy, federated meta-learning, rare disease prediction.

## I. INTRODUCTION

Accurate clinical prediction models can help the clinical decision makers identify the potential risk at an early stage so that appropriate actions can be taken in time in the pathway of care delivery. However, traditional disease prediction models are incapable of ensuring prediction accuracy with only a small amount of medical data, such as in the case of a rare disease. This problem is significant as there are about 7,000 rare diseases afflicting millions of people worldwide [1]. Furthermore, because of data privacy, hospitals and patients are still reluctant to share their health information. This leaves only a small sample of such data available for potentially predicting such rare diseases under the constraint that each hospital stores its diagnostic records at its own data vault (server) as shown in Fig.1. This scarce and constrained environment makes it difficult to predict rare diseases using neural networks which requires a lot more data for training.

Manuscript received August 16, 2021. This work was supported by the National Natural Science Foundation of China (62072469), National Key R&D Program (2018yfe0116700), Shandong Natural Science Foundation (ZR2019MF049) and the China Scholarship Council, (Corresponding author: Weishan Zhang)

Bingyang Chen, Tao Chen, Zhaoxiang Hou，Weishan Zhang and Xingjie Zeng are with the School of Computer Science and Technology, China University of Petroleum, Qingdao 266580, China (bychen@s.upc.edu.cn; zhangws@upc.edu.cn).

Jiehan Zhou is with the Oulu University.
Sumi Helal is with the University of Florida (helal@acm.org)

There are two main approaches of disease prediction: 1) time-series regression prediction [2-4] and 2) classifier-based prediction [5-7]. The first approach models and analyses the temporal dependencies between medical features to predict the risk probability of a disease. Time-series regression prediction mainly maps medical features into N-dimensional vectors by leveraging the encoding in natural language processing. Then a time series network is utilized to compute corresponding hidden states from each input medical feature to make predictions [2]. Classifier-based prediction, however, trains a classifier by a supervised classification system trained on medical datasets. However, disease prediction often requires large amounts of medical data to train such models. A data feature fusion approach [5] is proposed to improve heart disease categories prediction, which enriches data features by fusing electronic medical data with sensor data. This approach uses only multi-source data fusion in a single hospital to predict diseases, which does not fundamentally address the problem of multiple small samples and data privacy. In addition, the imbalance in disease categories is common in most medical scenarios [6]. This leads to low accuracy prediction results of rare diseases.

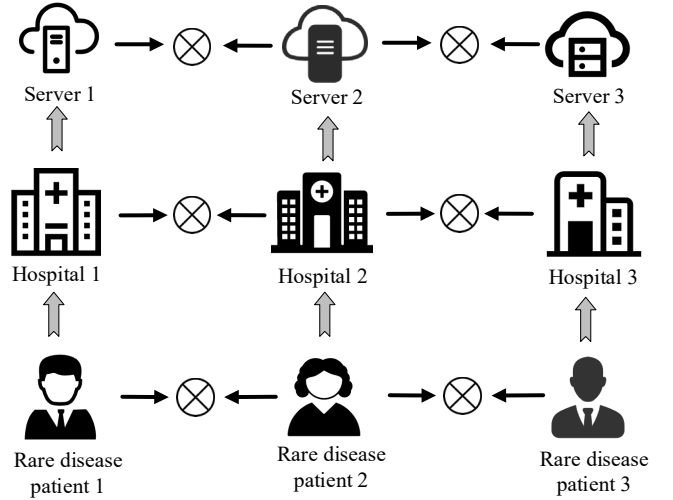

Fig. 1. Class imbalance and small samples in rare disease prediction

In short, there are two main challenges for machine learning approaches when applied to rare disease prediction:

1) it is difficult to ensure the accuracy and generalizability of the model for unseen diseases, due to class imbalance.

2) lack of sufficient medical data given the rarity of the diseases and the privacy concerns which makes it very hard to train models.

The complexity of medical data negatively affects disease identification. However, human physicians can recognize the rare disease through clinical experience. A model can also distinguish unseen classes if it is capable of "learning to learn", which is known as meta-learning. Specifically, MAML (Model Agnostic Meta-Learning) [8] algorithms can train the network with small samples of common categories to recognize other unseen categories. Using the method of modal movement rather than data movement, federated learning [9] realizes data fusion while protecting information security and privacy to further improve model performance. Integrating meta-learning and federated learning, results in a federated meta-learning framework [10] which effectively improves the model's classification accuracy of unseen classes.

To address all these challenges, and inspired by the framework in [10] described above, we utilize federated meta-learning to predict rare diseases, and apply two additional techniques to bolster its performance. We propose a novel dynamic-weight based fusion strategy and an attention mechanism based on federated meta-learning (DWA-FML) for rare disease prediction. In particular, we propose an attention-based meta-learning (ATML) approach based on the MAML architecture, which pays more attention to tasks with weaker training effects for improving identification results of unseen diseases. Moreover, to achieve effective data fusion while preserving data privacy, we designed a dynamic-weight based fusion strategy within federated learning to enable the meta-learning model to learn more categories of common disease characteristics to further improve the prediction accuracy of rare diseases. Specifically, each hospital (client) evaluates its local model and uploads it to a local server only when the newly evaluated model performs better than the previous version of the global model. After local models are uploaded to each hospital's local server, a central server, external to the hospitals, accesses the local servers to calculate the accuracy of each local model to use as the weight to update the global model. Data privacy is completely preserved in this approach where sensitive local medical data are not exposed or made accessible to the central server. Through experimentation, results show that our proposed approach achieves higher accuracy in less time (requiring only a few shots) than that of the original federated meta-learning approach. In fact, ATML achieves significantly higher performance than that of other meta-learning models on both medical image dataset as well as our dataset.

The primary contributions of this paper are as follows:

1) A novel machine learning approach for rare disease prediction. To the best of our knowledge, this is the first work to study rare disease prediction based on federated meta-learning.

2) An attention-based meta-learning approach to improve the prediction model performance for rare diseases. Our approach, as we show through experimental evaluations, performs better than that of other meta-learning approaches.

3) A dynamic-weight based fusion strategy to enable the privacy-preserving federation of the meta-learning architecture, which meets participating hospitals' data privacy policies.

4) A comprehensive and detailed evaluation of the proposed approach using rare disease datasets from actual hospitals.

Results show that our federated meta-learning model outperform the original federated meta-learning in terms of accuracy and speed of prediction.

The remainder of the paper is organized as follows. Section II discusses related work. Section III presents the approach for rare disease prediction in detail. Section IV evaluates the approach. Section V concludes the paper and points out important future work.

## II. RELATED WORK

**Meta-learning for disease prediction.** Meta-learning was proposed to address the weak adaptation of traditional neural networks to new tasks [11], which includes metric learning [12], [13], and model-agnostic meta-learning (MAML) [8]. The former had been utilized to learn the kernel-based regularization approach for blood glucose prediction [14]. The second approach can identify unseen classes with only a few samples of common classes and has shown promising results [15], which may be applied in medical data analysis. Specifically, in training, a support set consisting of a small number of samples from different classes is used to train the network. Similarly, a query set is utilized to evaluate the loss of the support set. During the test, the meta-learner can quickly adapt to new tasks with few fine-tuning steps. Due to limited medical samples, meta-learning has been employed for clinical risk prediction [7]. It proposes a model-agnostic gradient descent framework to train a meta-learner with a set of prediction tasks where the target clinical risks are highly relevant. However, on the one hand, we are usually not aware of potential relationships between source and target domains. On the other hand, there are little data in the source domain and even we cannot find the source domain. The ability to identify unseen classes is also used for rare skin disease diagnosis [16], which utilize image argumentation techniques and a difficulty-aware meta-learning algorithm to scale the loss for improving disease classification performance. The idea of a difficulty-aware meta-learning algorithm is worth learning and can be further improved.

**Federated learning for disease diagnosis.** Federated learning is a mechanism of training a shared global model on a central server, using only updates of local models stored and maintained in a federation of local sites such as hospitals or other institutions. Sensitive data in local institutions are used to perform model learning [17] but are not accessible by or visible to the global model and its central server. Specifically, each hospital (client) uses its own clinical data to train its local model without any need for data sharing between hospitals. Federated learning has in fact become the most effective approach to privacy-preserving data fusion in health care [18]. The trained local models are variably aggregated in a central server according to each hospital's contribution to the global model update, which is based on the size of data used in updating the hospital's local model [9]. The updated model is then dispatched to each client for the next round of training. A federated uncertainty-aware learning method is proposed to improve preterm birth prediction results, which reduces the contribution of models with high uncertainty in the aggregated model [19], but the best uncertainty evaluation criterion is difficult to determine. A dynamic fusion strategy is applied in federated learning for COVID-19 detection [20]. The client that performs better than the previous version is selected for average fusion in each episode. However, the average fusion method may not be the most effective fusion approach.

**Federated Meta-Learning.** The framework of federated meta-learning is first proposed by Huawei in 2018 [10], which focuses on improving the transmitting parameterized algorithm between mobile devices and central servers for faster convergence and higher accuracy. Different from federated learning, each client uses the parameters distributed by the server to initialize and train the local model. Local models of different clients are uploaded to a central server for updating the global model. Considering federated learning only focuses on a common output for all the users, MIT [21] proposes a theoretical understating regarding federated meta-learning for personalized study. Due to data privacy and scarce of credit frauds samples, federated meta-learning has been used to detect credit card fraud with strong detection performance [22]. However, the metric learning effects of federated meta-learning may have discrepancies in different tasks.

## III. APPROACH

In this section, we elaborate on the proposed federated meta-learning-based approach for rare disease prediction. We first present attention-based meta-learning (ATML) for rare disease identification. Then we describe how a dynamic-weight based fusion strategy integrates ATML model in federated meta-learning. Each client uses the meta-learning approach to train models, thus local meta-models are used in the following description to refer to local models. The system model of the proposed federated meta-learning approach is shown in Fig.2.

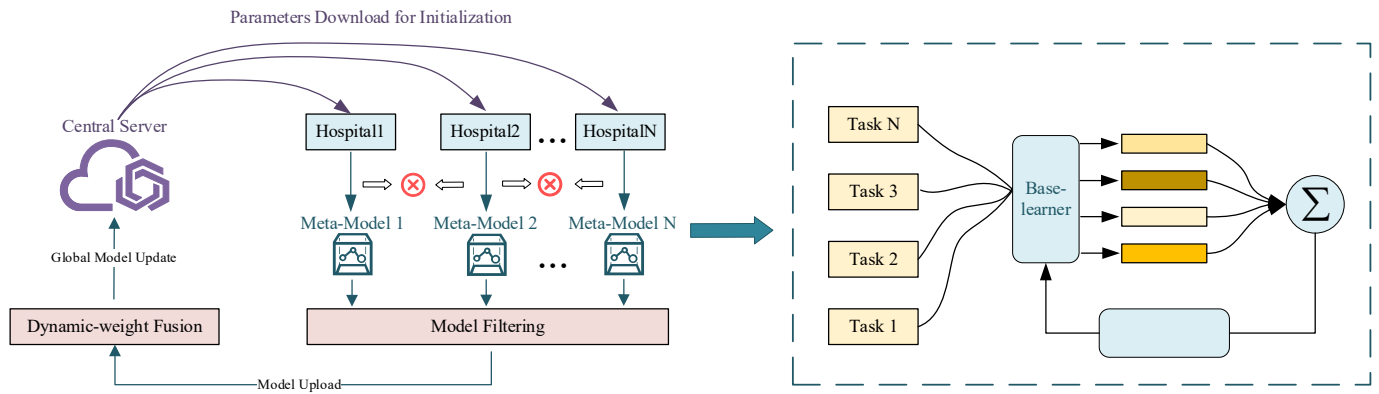

Fig. 2. Federated meta-learning-based approach for rare disease prediction (Each hospital means each client (for meta-learning), and meta-model means local meta-model (for federated meta-learning))

TABLE I
PROBLEMS AND SOLUTIONS

|  | Small samples | Classes imbalance | Identify unseen classes | Data privacy |
|---|---|---|---|---|
| Federated learning | √ |  |  | √ |
| Meta-learning | √ | √ | √ |  |

### A. Overview

TABLE I summarize the strong points of both federated learning and meta-learning, and it is obvious that combining both of them is very attractive to resolve the previous mentioned problems.

In federated disease prediction, suppose there are $n$ hospitals $h_i (i = 1,2,...,n)$ as clients. At each iteration step $e_i (i = 1,2,...,n)$ on the server, the local meta-models $f_{\theta^h}$ from each hospital are aggregated to update the global model $F_\theta$. Let the accuracy of the global model and the local meta-model on the test data $D^{te}$ be $Acc_s^{e_i}$ and $Acc_{h_i}^{e_i}$ respectively in the $e_i$th iteration step. The learning rate is defined as $\gamma$ to control the training speed of a model.

Each hospital $h_i$ employs the meta-learning approach for rare disease prediction on their own private medical dataset $D_i = \{x_i^h, y_i^h\}(h = 1,2,...n)$. Where $x_i^h$ is disease sample and $y_i^h$ is the corresponding label. In the meta-model training, tasks are randomly sampled in the set of $D^{tr}$ common diseases, consisting of a support set $D_S^{tr}$ and a query set $D_Q^{tr}$. Each local meta-model is initialized with parameters $\theta$ from the server then trained with $D^{tr}$. Loss $L(f_\theta(x_i^h), y_i^h)$ is evaluated by the error of base-model $f_\theta$ (base-learner in meta-learning). The learning rate of the inner loop is defined as $\alpha$ and outer loop as $\beta$. In meta-testing, rare diseases as test sets $D^{te}$, also consisting of support sets $D_S^{te}$ and query sets $D_Q^{te}$. The support set $D_S^{te}$ is utilized for fine-tuning the trained local meta-model $f_{\theta^h}$ to assess its performance on the query set $D_Q^{te}$.

Hence, we exploit federated meta-learning to predict the unseen disease with the learning experience of common diseases, which can improve model classification accuracy while protecting data privacy. The specific process of our approach is shown in Algorithm 1, which can be summarized in seven steps.

Step1: In federated learning, at each iteration step, the local meta-model(hospital) with better accuracy than the global model(server) of the previous round is dynamically selected and uploaded to the server.

Step2: During model fusion, ***dynamic-weight*** fusion is performed according to the accuracy of the uploaded local meta-model to update the global model. (line 9-11).

Step3: The parameters are downloaded to each hospital from the central server for model initialization to train specific tasks.

Step4: Instead of cross entropy loss, the focal loss is utilized for evaluating each task error in the inner loop.

Step5: **AT_Loss** are employed for updating parameters in the outer loop (line 20-21).

Step6: The trained local meta-model of each hospital is prepared for uploading to the central server.

Step7: Return to the first step until model convergence.

**Algorithm 1:** DWA-FML for rare disease prediction

1  // Run on Server
2  **Global Model Update**
3  Initialize $\theta$ for each hospital $h_i$
4  Retrieve $f_{\theta^{h_i}}$ from each hospital $h_i (i = 1,2 ...)$
5  **for** each episode $E = e_1, e_2 ... e_i$ **do**
6      $Acc_s^{e_{i-1}} \leftarrow$ global model evaluating in $D^{te}$
7      $Acc_{h_i}^{e_i} \leftarrow$ local meta-model $f_{\theta^{h_i}}$ test on $D^{te}$
8      **if** $Acc_s^{e_{i-1}} \leq Acc_{h_i}^{e_i}$ **then**
9          $Acc_{h_j \ (h_j \in h_i)}^{e_i} = Acc_{h_i}^{e_i}$
10         $w^{e_i}_k \leftarrow Acc_{h_k(h_k \in h_j)}^{e_i} / \sum_{j=1}^{n} Acc_{h_j}^{e_i}$
11         Global model $F^{e_i}_\theta \leftarrow \sum_{k=1}^{n} w_k \times f_{\theta^{h_k}}$
12     **end if**
13 **end for**
14 // Run on each hospital $h_i$
15 **Meta-model Update**
16 Sample support set $D_S^{tr}$ and query set $D_Q^{tr}$
17 $f_\theta \leftarrow$ download $\theta$ from server
18 $L_{D_S^{tr}}(\theta) \leftarrow \frac{1}{D_S^{tr}} \sum_{(x_i^h, y_i^h) \in D_S^{tr}} Focal\_Loss(f_\theta(x_i^h), y_i^h)$
19 $\theta^h \leftarrow \theta - \alpha \nabla_\theta L_{T_i}(f_\theta)$
20 $L_{D_Q^{tr}}(\theta^h) \leftarrow \frac{1}{D_Q^{tr}} \sum_{(x_i^h, y_i^h) \in D_Q^{tr}} AT\_Loss(f_{\theta^h}(x_i^h), y_i^h)$
21 $\theta^{h'} \leftarrow \theta^h - \beta \nabla_\theta L_{D_S^{tr}}(\theta^h)$
22 Return $f_{\theta^{h'}}$ to server

*B. Attention-based meta-learning for disease prediction*

MAML [8] model pays equal attention to all tasks. However, in practice, it is common that good classification accuracy can be achieved for simple tasks while poor accuracy for difficult tasks. To address this challenge, in each hospital $h_i$, we propose an attention-based meta-learning approach as shown in Algorithm 1. A series of tasks $T$ are randomly sampled in each episode with a task distribution $p(T)$ manner, which consist of support set $D_S^{tr}$ and query set $D_Q^{tr}$. The initialization parameters $\theta$ is dispatched to each hospital(client) by the server. Then $\theta$ is updated to $\theta^h$ by gradient descent on support set $D_S^{tr}$ in "adaption steps" (inner loop). We used focal loss [23] $Focal\_Loss(f_\theta(x_i^h), y_i^h)$ to improve the learning ability of the model. The focal loss function (Eq.1, Eq.2) and gradient descent (Eq.3) are defined as:

$$Ce\_loss = Cross\_entropy(out, target) \quad (1)$$

$$Focal\_loss_i = \eta * (1 - e^{-Ce\_loss})^\lambda * Ce\_loss \quad (2)$$

$$\theta^h \leftarrow \theta - \alpha \nabla_\theta Focal\_loss_{T_i}(f_{\theta^h}) \quad (3)$$

Where $\eta$ is a balanced variant of focal loss, $\lambda$ is focusing parameter satisfying $\lambda \geq 0$. $f_{\theta^h}$ denotes the base-model trained on $D_S^{tr}$.

In the "evaluation steps" (outer loop), the base-model $f_{\theta^h}$ is evaluated on query set $D_Q^{tr}$. Accordingly, the loss $L_{D_Q^{tr}}$ is computed to update the base-model. We propose that lower accuracy tasks (complex tasks) should contribute more to losses, an $AT\_Loss$ function is designed to further improve the accuracy of rare disease prediction. The attention-based meta optimization function instantiated as follows:

$$AT\_Loss = \sum_{T_i \sim p(T)} -Focal\_loss_i^\varphi * \log_2 accuracy_i \quad (4)$$

$$\theta^{h'} \leftarrow \theta^h - \beta \nabla_\theta AT\_Loss \quad (5)$$

Where $\varphi$ is a scaling factor to regulate the model's attention to difficult tasks.

We utilize Transformer (two layers are used) as the main network architecture of ATML (attention-based meta-learning). Specifically, the encoder of Transformer is employed for feature extraction to a m-dimensional vector $M$. To alleviate the negative impact of excessive differences between features, batch normalization is utilized to convert original data into b-dimension vector $B$. Then the vector $M$ and $B$ are concatenated into a $m + b$ dimensional vector to feed into the fully connected layer. As fewer training classes available will lead to weak sample diversity, we apply an Adam optimizer and set weight decay to 0.1 to improve the test performance of meta-model. The learning rate is defined as 0.001 to perform disease classification.

*C. The Federated Meta-Learning-based Framework*

In federated meta-learning, we aim that each hospital (client) can "share" their private dataset to design an effective rare disease prediction solution by increasing the categories of common diseases. Each hospital $h_i$ downloads parameter $\theta$ from the server for initialization, and trains the model $f_\theta$ on support set $D_S^{tr}$, and consequently tests the base-model $f_{\theta^h}$ on the query set to update the parameters from $\theta^h$ to $\theta^{h'}$. After that it sends the meta-model $f_{\theta^{h'}}$ to the server. The server aggregates different local meta-models to update the global model and dispatch the novel initialization parameters for each hospital. A dynamic-weight based fusion strategy is proposed to improve the prediction result. In particular, all the local meta-models are uploaded and average fused at the first round of aggregation. In each of the following rounds, each local meta-model accuracy $Acc_{h_i}^{e_i}$ on the test set $D^{te}$ are compared with the global model $Acc_s^{e_{i-1}}$ of the previous round. If local meta-models perform better, they are selected to prepare for aggregation, otherwise, the corresponding hospitals are filtered out at this round. In addition, a fusion weight $w^{e_i}_k$ is utilized in the aggregation process. Specifically, $w^{e_i}_k$ is the ratio of the current model's accuracy $Acc_{h_k(h_k \in h_j)}^{e_i}$ to the sum of all selected local meta-models' accuracy $\sum_{j=1}^{n} Acc_{h_j}^{e_i}$. The details of rare disease prediction model training process are illustrated in Fig.2 and described in Algorithm 1.

IV. EVALUATION

In this section, we evaluate the performance of different models on Arrhythmia dataset [24]. First, we compare with other related models to show the effectiveness of the proposed approach. Then, we make ablation experiments to explore which part of the improvement contributes most to the predicted results. We will discuss each meta-model effects with different iteration steps.

TABLE II
STATISTICS OF SELECTED DATASET

| Code | Class | Number |
|---|---|---|
| 1 | Normal | 245 |
| 2 | Right bundle branch block | 50 |
| 3 | Ischemic changes | 44 |
| 4 | Sinus bradycardia | 25 |
| 5 | Others | 22 |
| 6 | Old Anterior Myocardial Infarction | 15 |
| 7 | Old Inferior Myocardial Infarction | 15 |
| 8 | Sinus tachycardia | 13 |
| 9 | Left bundle branch block | 9 |

*A. Dataset description*

Arrhythmia dataset [24] contains 279 dimensional features. We use nine classes of diseases as shown in Table II. Additionally, medical image dataset [25] is also used to demonstrate the effectiveness of our approach.

*B. Comparison Experiment*

*(a) Comparisons with existing meta-learning approaches*

We consider the first five classes with large samples as common diseases (meta-train dataset $D^{tr}$), and other classes as rare diseases (meta-test dataset $D^{te}$). Each task is randomly sampled as distribution $p(T)$ from $D^{tr}$ and $D^{te}$. In meta-training, each task $T_i$ is a binary classification consisting of two random classes with $K$ samples per class in $D^{tr}$. The query set samples are twice the support set samples. Experiments showed that the model converges when iterates 1300 rounds, so we set 1500 rounds as the number of iterations and applied the attention-based optimization approach. In particular, we set the parameters that $\eta=5$, $\lambda = 2$, $\varphi = 2$ via experiments. The number of tasks is 10 and iteration per adaption step is 5. Similarly, in the meta-testing, two classes are randomly selected from the meta-test dataset $D^{te}$, and K samples of each class are sampled. The final experimental results are the average accuracy over 30 runs. In addition, experiments are also performed using the medical image dataset to demonstrate $AT\_Loss$ effectiveness in the meta-learning, as evaluated with AUC like Li [16] did.

TABLE III
THE PERFORMANCE OF DIFFERENT MODAL ON BOTH ARRHYTHMIA AND SKIN LESION DATASET

| | Arrhythmia | | | Skin Lesion | | |
|---|---|---|---|---|---|---|
| | 1 shot | 3 shot | 5 shot | 1 shot | 3 shot | 5 shot |
| Relation Net[13] | 61.00% | 62.74% | 72.37% | 59.97% | 62.87% | 72.40% |
| MAML[8] | 64.17% | 77.78% | 80.68% | 63.77% | 77.98% | 81.20% |
| DAML [16] | 68.33% | 81.38% | 83.89% | 67.33% | 79.60% | 83.30% |
| ATML | 79.89% | 86.52% | 90.94% | 73.17% | 83.31% | 87.26% |

We compared some meta-learning models including Relation Net, MAML, DAML. As shown in TABLE III, in all sample settings, the proposed attention-based meta-learning approach has better performance than that of other three approaches on both datasets, especially for Relation Net. Due to the limited number of training classes, the metric-learning model cannot sufficiently extract diverse features, which is one of the most important reasons for its inferior results. In addition, our approach exploits the MAML architecture and improves the $DA\_Loss$ function of DAML, which indicates the effectiveness of attention-based meta optimization in rare disease prediction.

*(b) Comparison with other algorithms*

Some widely used algorithms are selected for comparison experiments. Since there were no fine-tuning processes, baseline models were directly trained on $K$ shots per rare disease classes. Both the LSTM and the Transformer have two layers, while MLP consists of four linear layers. The iteration steps are five and other settings as the same as the meta-learning in Section IV. As shown in TABLE IV, the Transformer has stronger representation capability of feature extraction and does achieve good performance. However, compared with ATML, these classical algorithms have no capability to identify the rare disease. In addition, ATML with one shot is much better than others with five-shot, indicating the good performance of our approach for rare disease prediction.

TABLE IV
ACCURACY COMPARISON WITH SOME BASELINES

| Model | 1 shot | 3 shot | 5 shot |
|---|---|---|---|
| LR | 60.42% | 64.24% | 66.08% |
| KNN | 59.72% | 64.29% | 67.78% |
| MLP | 62.52% | 64.58% | 67.62% |
| LSTM | 61.11% | 62.74% | 67.48% |
| Transformer | 70.83% | 73.61% | 76.67% |
| ATML | 79.89% | 86.52% | 90.94% |

*(c) Comparison experiments of federated meta-learning*

In practical scenarios, it is impossible for a hospital to possess all disease categories for model training. To simulate the actual rare disease predicted situation in each hospital, we randomly selected three of the original five common disease classes as a training support set. Specifically, each local meta-model is trained by the three classes of common diseases to predict four classes of rare disease, which is denoted as $ATML_3$ to make a fair comparison with DWA-FML. In addition, classical federated learning with average fusion is denoted as FedAvg for a comparison experiment, where the Transformer is utilized for feature extractor. Similarly, the integration of FedAvg with MAML is denoted as FedAvg (MAML), and the integration of FedAvg with $ATML_3$ is denoted as FedAvg ($ATML_3$). As shown in TABLE V, there are different environments for one central server and four clients. The model training samples are five-shot per class.

TABLE V
EXPERIMENT ENVIRONMENT

| | GPU | RAM | Python | CUDA |
|---|---|---|---|---|
| Server | RTX3090 | 24G | 3.8.10 | 11.2 |
| Client 1 | RTX 3090 | 24G | 3.8.10 | 11.2 |
| Client 2 | RTX 3090 | 24G | 3.8.10 | 11.2 |
| Client 3 | RTX 3080 | 10G | 3.8.10 | 11.2 |
| Client 4 | GTX1080Ti | 11G | 3.8.10 | 11.2 |

As shown in TABLE VI, DWA-FML achieves a significantly higher accuracy than that of other methods. FedAvg is trained with five classes, however, its results are even worse than that of $ATML_3$ trained with three classes. The reason is that the Transformer has no capability to identify unseen classes. The results of ATML are much better than that of MAML in the experiment of section IV B(a), while the performance of FedAvg(MAML) is a little better than that of $ATML_3$. The comparison result shows it is essential for the meta-model to be trained in more common disease categories. Compared with FedAvg($ATML_3$), DWA-FML is more effective using the dynamic-weight based fusion strategy.

TABLE VI
THE ACCURACY OF DIFFERENT HOSPITAL

| Model | Hospital1 | Hospital2 | Hospital3 | Hospital4 |
|---|---|---|---|---|
| $ATML_3$ | 76.94% | 74.44% | 73.05% | 75.38% |
| FedAvg | 75.27% | 73.33% | 72.72% | 74.16% |
| FedAvg (MAML) | 82.22% | 83.88% | 81.66% | 82.50% |
| FedAvg ($ATML_3$) | 86.11% | 86.38% | 85.55% | 88.61% |
| DWA-FML | 88.61% | 87.66% | 86.94% | 89.72% |

*C Ablation Experiment*

*(a) Model effects with fusion rounds*

We now explore the impact of relevant parameters on the DWA-FML. Fig. 3 shows the local meta-model prediction results of each hospital with fusion rounds. The accuracy continues to rise for the first 150 steps, then it becomes gradually stable. Since each meta-model needs to be iterated for five rounds locally, our model only needs 750 rounds to complete the convergence. Compared with the meta-learning method (without federated learning) that converged in 1300 rounds, our model is able to learn faster. It indicates the proposed federated learning improves the prediction accuracy of rare diseases and accelerates model training. In addition, all hospitals achieve the best results almost simultaneously, which further indicates the effectiveness and generalizability of our approach.

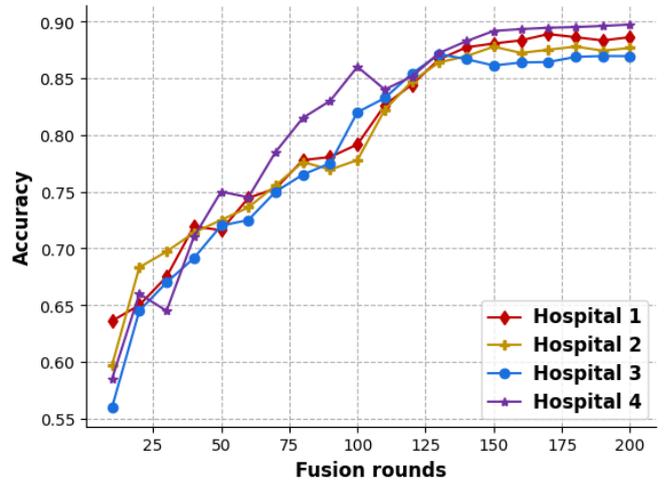

Fig. 3. Prediction results for different hospitals

*(b) Network architecture exploration*

In the ablation experiments of network architecture, we investigate the contribution of different network architectures to the proposed model. Specifically, FedAcc is the model that integrates FedAvg with the proposed dynamic-weight based fusion strategy, and feature extractor is the Transformer encoder. Similarly, FedAcc(MAML) means that MAML architecture is used in FedAcc.

TABLE VII
ABLATION EXPERIMENT ON NETWORK ARCHITECTURES

| Model | Hos1 | Hos2 | Hos3 | Hos4 | Avg |
|---|---|---|---|---|---|
| $ATML_3$ | 76.94 | 74.44 | 73.05 | 75.38 | 74.95 |
| FedAcc | 75.89 | 76.38 | 76.22 | 75.28 | 75.94 |
| FedAvg (MAML) | 82.22 | 83.88 | 81.66 | 82.50 | 82.57 |
| FedAvg ($ATML_3$) | 86.11 | 86.38 | 85.55 | 88.61 | 86.66 |
| FedAcc (MAML) | 84.44 | 84.16 | 83.05 | 87.22 | 84.71 |

| DWA-FML | 88.61 | 87.66 | 86.94 | 89.72 | 88.23 |

DWA-FML is the proposed model, which means FedAcc(ATML$_3$). Each hospital is denoted as Hos and the average accuracy of each hospital is denoted as Avg. Further, each result is a percentage.

As shown in TABLE VII, the performance of FedAcc is slightly better than that of ATML$_3$. This indicates the effectiveness of our fusion strategy. We further explore the effects of fusion strategy on DWA-FML. The accuracy of FedAcc(MAML) is 2.14% higher than that of FedAvg(MAML), and DWA-FML is 1.57% higher than that of FedAvg(ATML$_3$). The accuracy of FedAvg(ATML$_3$) is 4.09% higher than that of FedAvg(MAML) and DWA-FML is 3.52% higher than that of FedAcc(MAML). Therefore, the fusion strategy improves the prediction performance, the $AT\_Loss$ function is considered make a more contribution to the proposed approach.

### D  Further discussion on model performance and tome consumption

*(a)  Model performance with fine-tuning update steps*

Due to the necessity of fast adaption to new tasks, we set fine-tuning steps from one to ten in our experiments. As shown in Fig. 4, the performance increases rapidly in the first five fine-tuning steps and gradually levels in the next steps. The experiment results further illustrate our approach has good potential for rare disease prediction.

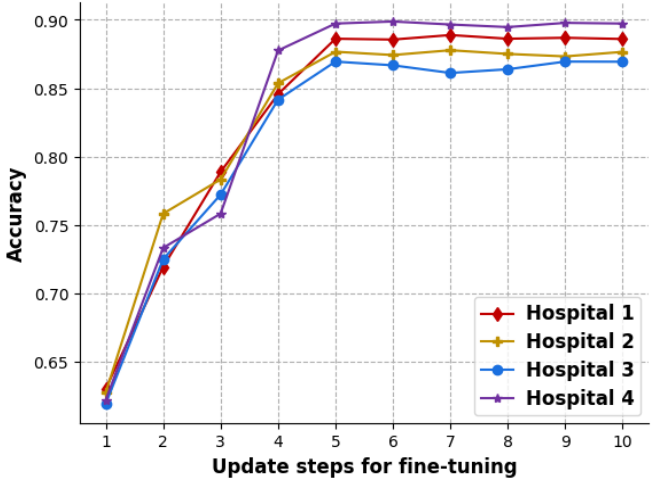

Fig. 4. Prediction results with fine-tuning steps

*(b)  Model performance with training steps*

Fig. 5 shows that most hospitals achieve the best results at the fifth training step. However, the models' performance degrades in the next training steps due to multiple training on common disease may lead to model over-fitting. In addition, compared with other clients, Hospital2 requires more training steps (eight steps) to achieve the best results, implying that the prediction performance of the meta-model in Hospital2 is weak. Furthermore, we find that the Hospital4's prediction accuracy is higher than that of other hospitals in both the fine-tuning (Fig. 4) and training process, indicating its meta-model is easier to train and has strong stability. Due to the use of dynamic-weight based fusion strategy, we believe that the local meta-model of the fourth hospital contributes more to the model aggregation of the server.

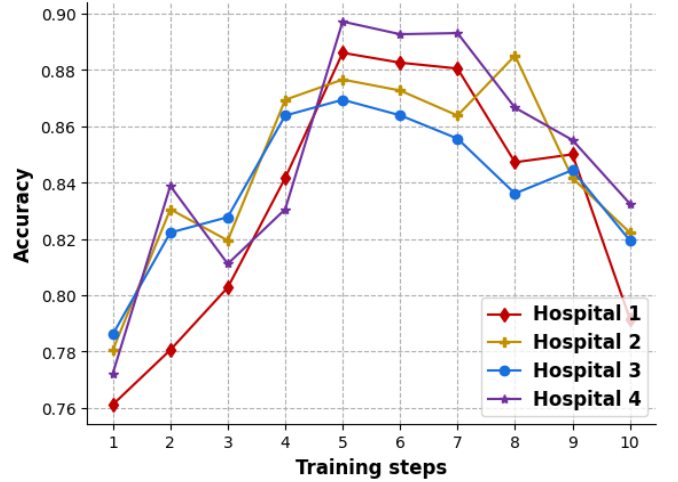

Fig. 5. Prediction results with training steps

*(c)  Time consumption*

This section discusses the time consumption of different models. Federated learning integrates different client models to achieve comparison experiments.

TABLE VIII
TIME CONSUMPTION

| Model | Time Consumption (min) |
|---|---|
| FedAvg(Transformer) | 16.43 |
| FedAcc(Transformer) | 16.22 |
| FedAvg(MAML) | 45.05 |
| FedAcc(MAML) | 40.12 |
| FedAvg(ATML) | 44.31 |
| DWA-FML | 39.91 |

DWA-FML is the proposed model, which means FedAcc(ATML)

As shown in Table VIII, the time consumption of federated learning model is much less than federated meta-learning. Because each local model (Transformer) is directly trained on rare diseases and then used to predict rare diseases. This process can be seen as the testing part of meta-learning. Due to only the difference in loss function, ATML and MAML have similar time consumption, which is in line with our expectations. In addition, our proposed fusion strategy reduced the time consumption by comparing FedAcc(MAML) and FedAvg(MAML), DWA-FML and FedAvg(MAML) respectively. Each local meta-model is only uploaded when it performs better than the global model of the previous version. The reduced time consumption is the model upload time. The small differences in time consumption between FedAvg(Transformer) and FedAcc(Transformer) indicate the global model's accuracy is too weak, and the local model is uploaded almost every round. In general, the proposed approach reduces the time consumption significantly compared with that of original federated meta-learning.

## V. CONCLUSION AND FUTURE WORK

This paper proposes a novel and effective rare disease prediction approach based on federated meta-learning. First, we present a novel federated meta-learning-based approach for rare disease prediction. Second, we propose an attention-based meta-learning approach for enhancing the model attention to difficult tasks. Third, we design a dynamic-weight based fusion strategy for each client to decide the participation of the local meta-model according to its performance. The evaluation results show that the proposed approach achieves higher classification accuracy across hospitals with good performance compared to the counter parts. In addition, our model achieves better performance and less time consumption than that of the original federated meta-learning.

In future work, we will explore the communication efficiency of our approach and consider investigating how to use blockchain technology to further improve data security and privacy.